\documentclass[sigconf,screen]{acmart}

\AtBeginDocument{%
  }

\copyrightyear{2025}
\acmYear{2025}
\setcopyright{acmlicensed}
\acmConference[MM '25]{Proceedings of the 33rd ACM International Conference on Multimedia}{October 27--31, 2025}{Dublin, Ireland}
\acmBooktitle{Proceedings of the 33rd ACM International Conference on Multimedia (MM '25), October 27--31, 2025, Dublin, Ireland}
\acmDOI{10.1145/3746027.3755200}
\acmISBN{979-8-4007-2035-2/2025/10}

\usepackage{amsmath,amsfonts}
\usepackage{algorithm}
\usepackage{algpseudocode}
\usepackage{textcomp}
\usepackage{stfloats}
\usepackage{url}
\usepackage{verbatim}
\usepackage{graphicx}
\usepackage{threeparttable}
\usepackage{makecell}
\usepackage{multirow}
\usepackage{footnote}
\usepackage{tablefootnote}
\usepackage{float}
\usepackage{colortbl}
\usepackage{array}
\usepackage{caption}
\usepackage{booktabs}
\usepackage{ragged2e}
\usepackage{setspace}
\usepackage{enumitem}
\usepackage{bbm}
\usepackage{subcaption}
\usepackage{float}
\usepackage{balance}

\makeatletter
\def\whline#1{%
	\noalign{\ifnum0=`}\fi\hrule \@height #1 \futurelet
	\reserved@a\@xhline}

\usepackage{paralist}

\definecolor{seagreen}{rgb}{0.1,0.92,0.21}
\definecolor{darkgreen}{rgb}{0.21,0.82,0.11}
\definecolor{deepblue}{rgb}{0.12,0.25,0.62}
\definecolor{darkred}{rgb}{0.92,0,0.12}
\definecolor{lightred}{RGB}{255,235,235}

\begin{document}

\title{Capturing More: Learning Multi-Domain Representations for Robust Online Handwriting Verification}

\author{Peirong Zhang}
\orcid{0000-0002-1857-5473}
\affiliation{
	\institution{South China University of Technology}
	\city{Guangzhou}
	\state{Guangdong}
	\country{China}
}
\email{eeprzhang@mail.scut.edu.cn}

\author{Kai Ding}
\authornote{Corresponding authors.}
\orcid{0000-0002-9371-0751}
\affiliation{
	\institution{INTSIG Information Co. Ltd}
	\institution{INTSIG-SCUT Joint Lab on Document Analysis and Recognition}
	\city{Shanghai}
	\country{China}
}
\email{danny\_ding@intsig.net}

\author{Lianwen Jin}
\authornotemark[1]
\orcid{0000-0002-5456-0957}
\affiliation{
	\institution{South China University of Technology}
	\institution{SCUT-Zhuhai Institute of Modern Industrial Innovation}
	\city{Guangzhou}
	\state{Guangdong}
	\country{China}
}
\email{eelwjin@scut.edu.cn}

\begin{abstract}
In this paper, we propose SPECTRUM, a temporal-frequency synergistic model that unlocks the untapped potential of multi-domain representation learning for online handwriting verification (OHV). SPECTRUM comprises three core components: (1) a multi-scale interactor that finely combines temporal and frequency features through dual-modal sequence interaction and multi-scale aggregation, (2) a self-gated fusion module that dynamically integrates global temporal and frequency features via self-driven balancing. These two components work synergistically to achieve micro-to-macro spectral-temporal integration. (3) A multi-domain distance-based verifier then utilizes both temporal and frequency representations to improve discrimination between genuine and forged handwriting, surpassing conventional temporal-only approaches. Extensive experiments demonstrate SPECTRUM's superior performance over existing OHV methods, underscoring the effectiveness of temporal-frequency multi-domain learning. Furthermore, we reveal that incorporating multiple handwritten biometrics fundamentally enhances the discriminative power of handwriting representations and facilitates verification. These findings not only validate the efficacy of multi-domain learning in OHV but also pave the way for future research in multi-domain approaches across both feature and biometric domains. Code is publicly available at \url{https://github.com/NiceRingNode/SPECTRUM}.
\end{abstract}



\maketitle

\section{Introduction}
Evolving from quill and ink to the digital age, handwriting verification has long been a fundamental technique for identity authentication. It plays crucial roles in diverse applications, such as financial transactions, legal proceedings, and government operations. Signatures have traditionally been the primary handwritten biometric for handwriting verification \cite{advoffline2019hafemann,deepsign2021tbiom,sig2vec2022lai}. Beyond signatures, recent efforts have started to explore more handwritten biometrics such as isolated digits \cite{biotouchpass2019tmc,biotouchpass22020tifs} or consecutive digit strings \cite{msds2022zhang,pavenet2025zhang}, enriching the versatility of the available handwritten mediums and broadening the utility of this field. Generally, handwriting verification can be categorized into two manners: online and offline \cite{onoffreview2012ferrer,review2019diaz}. Online techniques \cite{ran2019lai} utilize dynamic data produced in the writing process, such as speed and pressure, for authentication, whereas the offline counterpart \cite{advoffline2019hafemann} analyzes digitized handwritten images obtained by scanning or photographing. In this paper, we focus on online handwriting verification (OHV).

The key challenge of OHV lies in extracting features that effectively capture unique handwriting styles. Therefore, a wide range of feature modeling techniques have been explored, such as temporal features \cite{ran2019lai,deepsign2021tbiom,sig2vec2022lai,dsdtw2022jiang}, frequency features \cite{dwt2006ieice,dwt2008prl,wavelet2018osv}, and statistical features \cite{gmmhmm2005liang,pca2006osv,hmm2018farimani}. In recent years, temporal modeling has become the \emph{de facto} paradigm that dominates the state-of-the-art, primarily leveraging techniques like dynamic distance warping (DTW) or convolution/recurrent neural networks (CNN/RNN) to capture the temporal dynamics inherent in handwriting. Frequency features, typically derived from temporal features using Fourier or Wavelet transform, offer another powerful analytical tool for OHV. While once widely used, their application has been largely limited to superficial feature extraction \cite{dwt2006ieice,dwt2008prl}. This constrained utilization has hindered their potential, resulting in diminished academic interest in frequency-based approaches.

Current OHV methods predominantly rely on temporal features alone, potentially missing crucial signature characteristics that could enhance verification accuracy. Drawing insights from related fields, a multi-domain approach could address these limitations. For instance, face forgery detection \cite{f3net2020eccv,highfreqtf2023miao}, speaker verification \cite{mfa2022liu,golden2024liu}, and online writer retrieval \cite{dolphin2024zhang} utilize frequency subbands to enrich RGB images, audio signals, or online handwriting traits. They have achieved superior performance and demonstrate the value of multi-domain learning. However, despite this proven effectiveness in parallel domains, the potential of multi-domain feature learning remains largely unexplored in OHV.

\begin{figure*}
	\centering
	\includegraphics[width=0.82\textwidth]{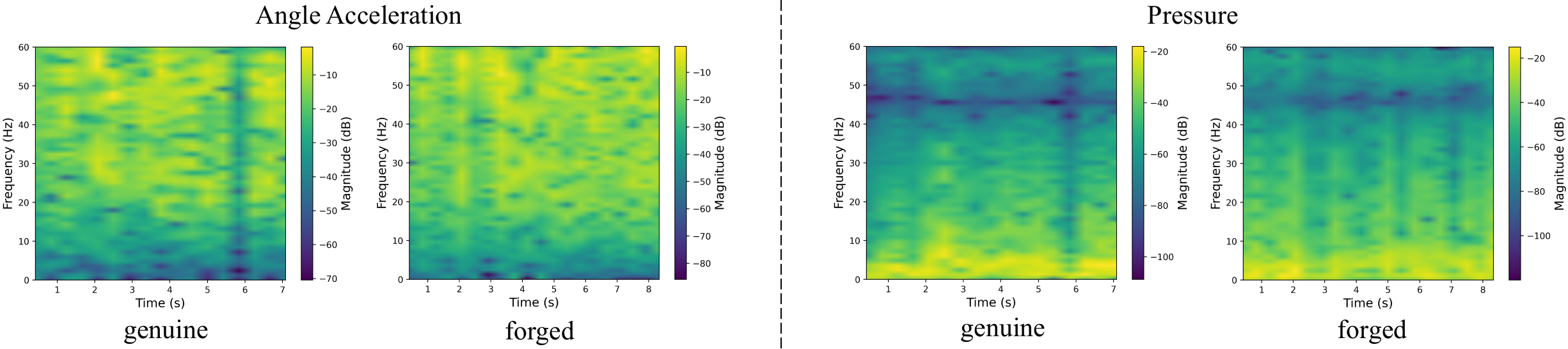}{}
	\caption{Spectrograms of time-domain features extracted by short-time Fourier transform (STFT) on genuine and forged handwriting samples, in which angular acceleration and pressure are taken as example features. The frequency responses of genuine and forged handwriting showcase obvious discrepancies. Hence, frequency modeling offers another discriminative perspective and can be combined with temporal features to achieve multi-domain discrimination.}
	\label{Fig::stft}
\end{figure*}

Given the intrinsic connection between temporal and frequency domains, and inspired by successful multi-domain approaches in related fields, we investigate whether frequency features could complement temporal learning in OHV. Using short-time Fourier transform (STFT), we extract spectrograms of time-domain signature features as demonstrated in Fig.~\ref{Fig::stft}. The results reveal significant discrepancies between genuine and forged handwritten signatures in the frequency domain, capturing unique writing characteristics such as rhythms and periodicities that temporal features might miss. Therefore, leveraging frequency features to enhance temporal analysis offers a natural and promising path toward multi-domain OHV, which potentially unlocks more discriminative handwriting representations and improves verification performance.

To this end, we propose \textbf{SPECTRUM}, a \textbf{SPEC}tral-\textbf{T}empo\textbf{R}al \textbf{U}nified \textbf{M}odel that integrates temporal and frequency domains for online handwriting verification. First, we design two components to achieve micro-to-macro multi-domain integration (\textbf{M$^3$I}). (1) \emph{Micro integration.} We propose a \textbf{multi-scale interactor} to facilitate fine-grained interaction between temporal and frequency features. At each scale, the handwriting sequence is split into even and odd sub-sequences for independent temporal and frequency processing. Temporal features are preserved via a projection layer, while frequency modeling is performed by combining the 1D (inverse) Fourier transform with learnable complex weights at scale $l$ to emphasize salient frequency features \cite{gfnet2023rao}. The two sub-sequences are then interleaved to promote mixed-domain interaction. Applying this procedure at multiple scales, this module enables the aggregation of diverse contextual cues and scale-wise complementarity. (2) \emph{Macro integration.} We introduce a \textbf{self-gated fusion module} that dynamically weights the contributions of global temporal and frequency features, enabling self-optimized feature fusion. Collectively, these two modules achieve comprehensive temporal-frequency interplay in a micro-to-macro manner. Second, we propose a multi-domain distance-based verifier (\textbf{MDV}) for inference optimization. MDV combines DTW distance computed by temporal features and Euclidean distance computed by frequency features during testing to enhance the discrimination between genuine and forged samples. By naturally harnessing representations of both domains, MDV transcends the reliance on merely temporal features in prior works, resulting in better verification accuracy.

We evaluate SPECTRUM using three online handwriting datasets: MSDS-ChS \cite{msds2022zhang} (Chinese Signature), MSDS-TDS \cite{msds2022zhang} (Token Digit String (TDS)), and DeepSignDB \cite{deepsign2021tbiom} (Latin Signature). Experiments demonstrate a pronounced outperformance of SPECTRUM over state-of-the-art OHV methods that solely depend on temporal representation learning. This evidences the effectiveness of the M$^3$I mechanism and MDV in incorporating frequency features for multi-domain learning. In addition, we investigate multi-domain fusion between multiple handwritten biometrics by combining Chinese signature and TDS to enrich individual writing representations. This approach further improves verification performance, suggesting that multi-domain learning can be extended across not only feature domains (temporal and frequency) but also biometric domains (Chinese signature and TDS) and potentially opens new avenues for future research.

Our main contributions are summarized as follows:
\begin{itemize}
	\item We propose SPECTRUM, a multi-domain representation model for online handwriting verification. By synergizing temporal and frequency information, SPECTRUM overcomes the limitations of traditional single-domain approaches, effectively enhancing signature representation quality.
	\item We design a multi-scale interactor and a self-gated fusion module inside SPECTRUM, effectively integrating multi-domain features in a micro-to-macro manner. In addition, we introduce a multi-domain distance-based verifier MDV, which naturally leverages both temporal and frequency representations and improves verification performance.
	\item Experiments demonstrate the superiority of SPECTRUM over existing OHV methods. We further reveal the effectiveness of incorporating multiple handwritten biometrics to enhance representation discrimination and OHV performance, potentially inspiring future research.
\end{itemize}

\section{Related Work}
\label{sec::related}
\subsection{General Online Handwriting Verification Techniques}
Online handwriting verification (OHV) has seen substantial progress in recent decades, primarily focusing on online signature verification \cite{review2019diaz} due to its pervasive usage. This technique typically constitutes two stages: feature representation and decision making. (1) \emph{Feature representation.} The evolution from traditional hand-crafted extraction methods \cite{gmmdtw2017sharma,gmm2017pr,tang2018info,okawa2021time} to modern deep learning methods has established new state-of-the-art performance. Current deep learning approaches broadly operate in two paradigms. The first type concentrates on local feature modeling, often developed in conjunction with Dynamic Time Warping (DTW). PSN \cite{wu2019prewarp} and TA-RNNs \cite{deepsign2021tbiom} pre-align handwriting sequences using DTW before inputting them to CNN/RNN-based models. DeepDTW \cite{wu2019DeepDTW} uses a DTW on top of a Siamese CNN to enhance local invariance learning. RAN \cite{ran2019lai} proposes a length-normalized path signature descriptor to describe local signature trajectories. DsDTW \cite{dsdtw2022jiang} integrates the differentiable soft-DTW into the loss function to improve local discriminative learning. The second paradigm captures global representations. Park et al. \cite{park2019robust} utilize an LSTM-CNN network to analyze features at both stroke and signature levels. Li et al. \cite{li2019stroke} progressively model the stroke features and the holistic signature with RNN. Sig2Vec \cite{sig2vec2022lai} proposes a selective pooling module, converging subspace features into a fixed-length vector with global context awareness. Xie et al. \cite{fbn2023xie} uses BERT \cite{bert2019devlin} as the backbone for global sequence modeling. (2) \emph{Decision making.} Typically configured in an open-set manner, OHV systems are trained on limited data but tested on unlimited unseen data. This requires models to generate feature vectors to assess similarities between templates and queries, thereby verifying queries' authenticity. Common approaches include Euclidean/DTW distance-based verifiers that authenticate queries falling within specific thresholds \cite{ran2019lai,sig2vec2022lai,dsdtw2022jiang}, subject-independent classifiers evaluating sample-wise distances \cite{wu2019DeepDTW,wu2019prewarp}, and sigmoid scoring based on pre-given thresholds \cite{deepsign2021tbiom}.

\begin{figure*}[t]
	\centering
	\includegraphics[width=\textwidth]{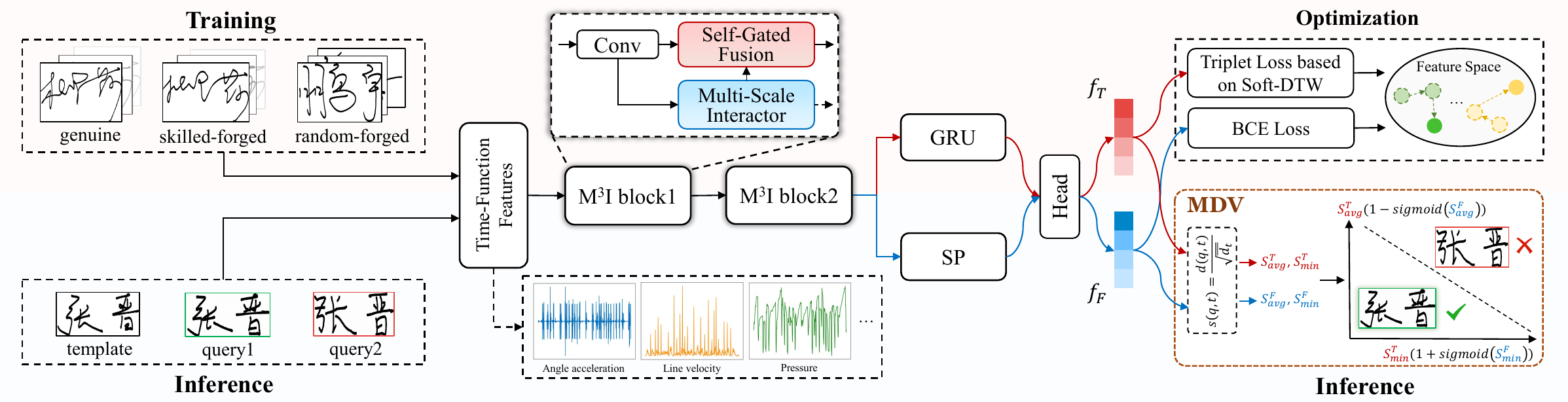}{}
	\caption{Overall framework of SPECTRUM. \textbf{Top}: Model training process. \textbf{Middle}: Detailed architecture of SPECTRUM, which mainly consists of two stacked micro-to-macro multi-domain integration (M$^3$I) blocks, a GRU, and a selective pooling (SP) layer \cite{sig2vec2022lai}. The last M$^3$I block exclusively outputs frequency features, which are pooled to yield $f_F$. \textbf{Bottom}: Model inference (verification) process, where MDV harnesses both temporal and frequency representations to enhance verification accuracy.}
	\label{Fig::overall}
\end{figure*}

Recently, the OHV field has included new handwritten biometrics like digit/digit strings beyond signatures. Tolosana et al. propose the e-BioDigit \cite{biotouchpass2019tmc} and MobileTouchDB \cite{biotouchpass22020tifs} datasets for second-level identity authentication using separate digits. Zhang et al. \cite{msds2022zhang} propose the MSDS dataset, including a novel MSDS-TDS subset that firstly leverages Token Digit String (TDS) as biometrics. They demonstrate that mainstream OSV methods can be seamlessly transferred to TDS verification, even achieving better performance than signature verification.

\subsection{Frequency Learning for Online Handwriting Verification}
While contemporary methods primarily rely on the temporal domain for handwriting analysis, earlier research has explored frequency analysis for handwriting characterization due to the natural bridge between temporal and frequency domains. The Wavelet transform \cite{dwt2006ieice,dwt2008prl,dwtnn2010asej,hmmdwt2017chavan,gmm2017yang,dctdwtdft2018miaba,wavelet2018osv} and Fourier transform \cite{fft2006quan,yanikoglu2009online,hmmdwt2017chavan,dctdwtdft2018miaba} are mostly adopted, while additional frequency features such as Discrete Cosine/Hartley/Walsh-Hadamard/Kreke/Mellin transform \cite{dwt2008prl,hmmdwt2017chavan,mfcc2011fallah} are also explored. Despite these efforts, frequency learning for OHV has been shackled by two critical drawbacks. (1) \emph{Limited feature extraction.} Most studies rely solely on frequency transforms for feature extraction without further modeling, usually yielding insufficient discriminative features. (2) \emph{Domain isolation.} Prior methods rely exclusively on the frequency domain but overlook the potential synergy with temporal modeling. This oversight also persists in current cutting-edge temporal-centric approaches, which solely focus on temporal modeling. To address these issues, we propose SPECTRUM, a multi-domain learning model that integrates temporal and frequency in a micro-to-macro manner, empowering handwriting representation from the multi-domain perspective.

\section{Methodology}
\label{sec::method}
Fig.~\ref{Fig::overall} illustrates the overall framework of the proposed SPECTRUM. Our model synergizes temporal and frequency domains through the multi-scale interactor and self-gated fusion module (Sec.~\ref{sec::m3i}), while using the multi-domain distance-based verifier (MDV) (Sec.~\ref{sec::mdv}) to enhance verification. The red paths of Fig.~\ref{Fig::overall} demonstrate the training process while the blue paths represent the inference process.

\subsection{Micro-to-Marco Multi-Domain Integration (M$^3$I) Mechanism}
\label{sec::m3i}
To fully combine temporal and frequency features, we propose a micro-to-macro multi-domain integration (M$^3$I) learning mechanism, which corresponds to the M$^3$I blocks depicted in Fig.~\ref{Fig::overall}.
\vspace{-5pt}

\begin{figure}[t]
	\centering
	\includegraphics[width=\linewidth]{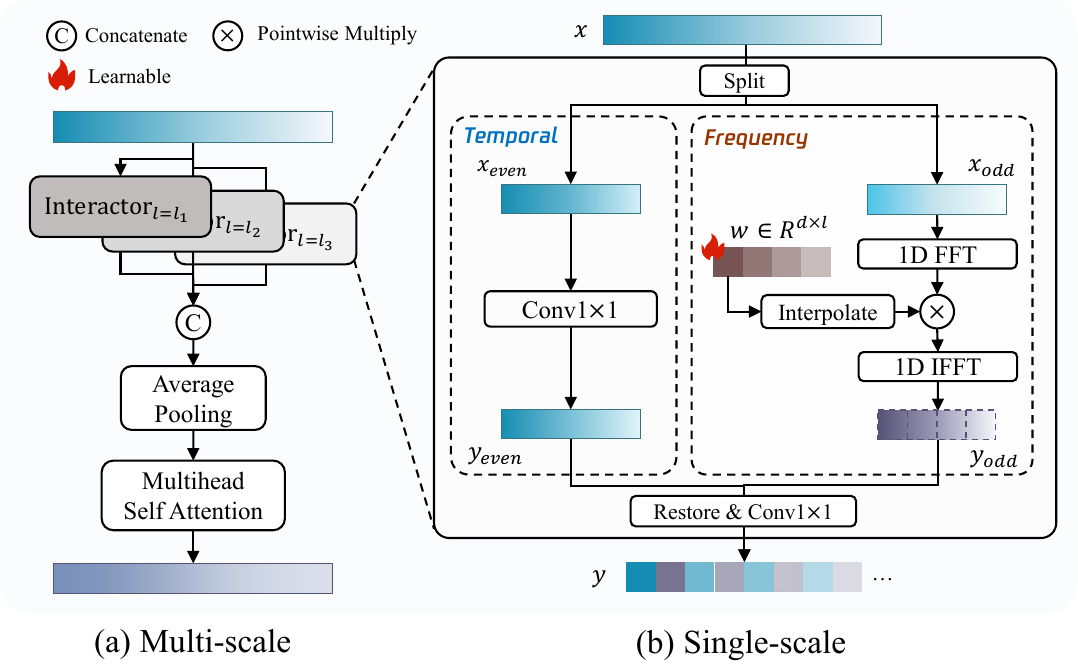}{}
	\caption{Schematic of the multi-scale interactor.}
	\label{Fig::multiscale}
\end{figure}

\paragraph{Micro-level multi-domain learning} We design a multi-scale interactor to capture fine-grained interactions between temporal and frequency features. As shown in Fig.~\ref{Fig::multiscale} (a), the multi-scale interactor is composed of multiple single-scale interactors, whose architecture is detailed in Fig.~\ref{Fig::multiscale} (b). We begin with illustrating the design of a single-scale interactor. Given an input temporal handwriting sequence $x \in \mathbb{R}^{d \times L}$ ($d$ is the embedding dimension and $L$ is the sequence length), we split it into two sub-sequences $x_{even} \in \mathbb{R}^{d \times \lceil L / 2 \rceil}$ and $x_{odd} \in \mathbb{R}^{d \times \lceil L / 2 \rceil}$ by separating even and odd timesteps along the spatial dimension. $x_{even}$ is dedicated to preserving temporal information and undergoes a simple $1 \times 1$ convolution to derive $y_{even}$. In contrast, $x_{odd}$ is assigned for frequency modeling. Inspired by \cite{gfnet2023rao}, we perform 1D discrete Fourier transform (DFT) on $x_{odd}$ to calculate its spectrum response $X$. Given each embedding dimension $i \in [0,d-1]_{\mathbb{Z}}$, the frequency response $X[i]$ for $x_{odd}[i]$ is calculated as:
\begin{equation}
	\label{Eq::fft}
	X[i,k] = \sum_{n=0}^{N-1} x_{odd}[i,n] e^{-j\frac{2\pi k}{N}n} \in \mathbb{R}^{1 \times N}, k \in [0,N - 1]_{\mathbb{Z}},
\end{equation}
\noindent where $N=\lceil L / 2\rceil$, $j$ is the imaginary unit, and $X[k]$ represents the frequency response of $x[n]$ at the frequency point $\omega_k=\frac{2\pi k}{N}$. By aggregating $X[i]$, we can obtain the entire frequency features $X=\{X[i]\} \in \mathbb{R}^{d \times N}$. For real-value inputs $x_{odd}[i,n]$, its DFT response is inherently symmetric \cite{conjugate1978dubois,gfnet2023rao}, \emph{i.e.}, $X[i,N-k] = X^*[i,k]$. Therefore, we retain only the first half of DFT for further processing, \emph{i.e.}, $\hat{X} = \{X[i,\hat{k}]\} \in \mathbb{R}^{d \times \lceil N/2\rceil}, \hat{k} \in [0,\lceil N/2 \rceil-1]_{\mathbb{Z}}$, which fully preserves the frequency characteristics of $x_{odd}$.

Subsequently, we introduce a 1D learnable complex weights $w \in \mathbb{R}^{d \times l}$ to modulate the frequency features $\hat{X}$ and selectively amplify the discriminative aspects. The length $l$ of $w$ reflects the ``scale'' term of the single-scale interactor. However, the predefined length $l$ in the model configuration may not match the spectral length $N$. Hence, we first interpolate $w$ to length $\lceil N/2 \rceil$ using bilinear interpolation, resulting in $\bar{w}$, and then multiply it with $\hat{X}$:
\begin{equation}
	\begin{gathered}
		\bar{w} = interpolate(w,\lceil N/2 \rceil),\\
		\bar{X} = \hat{X} \odot \bar{w},
	\end{gathered}
\end{equation}
where $\odot$ denotes point-wise multiplication. With the weighted frequency features, we perform 1D inverse Discrete Fourier transform (IDFT) on $\bar{X}[i]$ of each embedding dimension $i$. As $\bar{X}[i]$ represents the half-spectrum due to conjugate symmetry, we reconstruct the full-spectrum $\tilde{X}[i]$ and then perform the IDFT:
\begin{equation}
	\begin{gathered}
		\tilde{X}[i,k] = \begin{cases}
			\bar{X}[i,k], 0 \le k \textless \lceil N/2 \rceil,\\
			\bar{X}^*[i,N-k], \lceil N/2 \rceil \le k \textless N,
		\end{cases}\\
		y_{odd}[i,n] = \frac{1}{N}\sum^{N-1}_{k=0}\tilde{X}[i,k]e^{j\frac{2\pi k}{N}n} \in \mathbb{R}^{d\times N}.
	\end{gathered}
\end{equation}
Here, we derive the remapped output $y_{odd}$, representing frequency-modulated writing features. For efficient implementation, we adopt the functionally equivalent Fast Fourier transform (FFT) and inverse Fast Fourier transform (IFFT) to compute DFT and IDFT, reducing the computation complexity from $\mathcal{O}(N^2)$ to $\mathcal{O}(NlogN)$ and improving both training and inference efficiency.

Given the temporal output $y_{even}$ and frequency-modulated output $y_{odd}$, we restore them to a new sequence according to their original even and odd positions to intertwine the temporal and frequency features. The interleaved features are then passed through a $1\times 1$ convolution to derive the output $y$ of a single-scale interactor. Afterward, we build the multi-scale interactor using $m$ single-scale interactors with varying scales $l$s, feeding the input $x$ to each of them and consolidating their output by average pooling. We further impose a multi-head self-attention \cite{attention2017vaswani} on the averaged sequence and obtain the final mixed-domain output. $m$ is empirically set to 3. Ablation studies on the value of $m$ are detailed in Sec.~\ref{sec::ablation}.

\begin{figure}[t]
	\centering
	\includegraphics[width=0.5\linewidth]{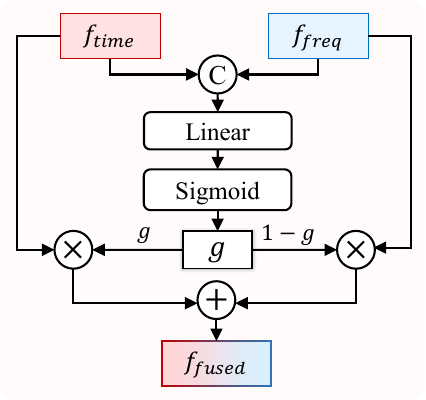}{}
	\caption{Schematic of the self-gated fusion module.}
	\label{Fig::gated}
\end{figure}

\vspace{-5pt}
\paragraph{Macro-level multi-domain learning} As shown in Fig.~\ref{Fig::overall}, the temporal features passed through the convolution module (the \emph{Conv} block) are fed into the multi-scale interactor for fine-grained temporal-frequency learning, outputting domain-interleaved features. More globally, these interleaved features can be further fused with the external temporal features. To this end, we introduce a self-gated fusion module for global multi-domain interaction as illustrated in Fig.~\ref{Fig::gated}. Given temporal features $f_{time} \in \mathbb{R}^{L\times d}$ and frequency-modulated features $f_{freq} \in \mathbb{R}^{L\times d}$, they are concatenated along the channel dimension to yield $f \in \mathbb{R}^{L \times 2d}$. We then compute a gate coefficient $g$ to dynamically fuse them:
\begin{equation}
	\label{Eq::gate}
	\begin{gathered}
		g = f \textrm{@} W^T + b, W \in \mathbb{R}^{d \times 2d}, b \in \mathbb{R}^{d},\\
		f_{fused} = f_{time} \odot g \oplus f_{freq} \odot (1-g),
	\end{gathered}
\end{equation}
where $\textrm{@}$ denotes matrix multiplication, $\odot$ signifies point-wise multiplication, $\oplus$ signifies point-wise addition, $W$ and $b$ are weights and biases of a linear layer. The fused features $f_{fused}$ are combined by adaptively weighting the contributions of temporal and frequency features through the self-derived gate $g$, accomplishing global temporal-frequency feature integration.

\vspace{-5pt}
\paragraph{Discussion} In the multi-scale interactor, the segmented sub-sequences $x_{even}$ and $x_{odd}$ retain much of the original sequence's dynamic and structural integrity despite the reduced resolution, ensuring sufficient fundamental handwriting characteristics for subsequent temporal and frequency feature extraction. Our frequency modeling approach follows \cite{gfnet2023rao}, but is tailored specifically for 1D handwriting sequences rather than 2D images. Through $x_{even}$'s transformation into the frequency domain and the modulation of learnable weights, our model adaptively emphasizes the unique writing patterns among specific frequency bands while filtering out noise. The following recombination naturally interweaves the temporal and frequency sequences, promoting deep interaction and complementarity between the two domains. Furthermore, the self-gated fusion facilitates a more holistic multi-domain consolidation with self-driven feature balance. These designs collaboratively enable a comprehensive micro-to-macro integration of temporal and frequency features.

\subsection{Multi-Domain Distance-Based Verifier}
\label{sec::mdv}
Similar to Sig2Vec \cite{sig2vec2022lai} and DsDTW \cite{dsdtw2022jiang}, SPECTRUM adopts an open-set OHV approach, enabling it to verify handwriting from unlimited writers previously unseen during training. Thus, it exploits a distance-based verifier that compares the feature representations of template and query handwriting for verification. Nevertheless, prior methods are confined to solely utilizing temporal embeddings in this process. Given the dual temporal and frequency awareness in SPECTRUM, we propose a multi-domain distance-based verifier (MDV) to leverage representations from both domains for enhanced discrimination. As shown in the right panel of Fig.~\ref{Fig::overall}, given two handwriting $x^i$ and $x^j$, they undergo model feature extraction $\phi$ and derive the temporal feature sequences $f_T^i,f_T^j \in \mathbb{R}^{L_T \times d}$ and frequency feature vectors $f_F^i,f_F^j \in \mathbb{R}^{d}$ ($L_T$ is the sequence length and $d$ is the embedding dimension). We compute the Dynamic Time Warping (DTW) distance between temporal sequences and Euclidean distance between frequency vectors as:
\begin{equation}
	\begin{gathered}
		d_T(x^i,x^j) = DTW(\phi(x^i),\phi(x^j)) = DTW(f_T^i,f_T^j) ,\\
		d_F(x^i,x^j) = ||\phi(x^i) - \phi(x^j)||^2 = ||f_F^i - f_F^j||^2.
	\end{gathered}
\end{equation}
Given $n$ template handwriting $\{x_u^1,...,x_u^n\}$ attributed to writer $u$, we compute average pairwise distance between their temporal features, denoted as $\bar{d_T^u}$ ($\bar{d_T^u} = 1$ if $n=1$). For a query handwriting $x^q$ claiming to be writer $u$, we compute temporal and frequency scores between $x^q$ and all templates:
\begin{equation}
	s_T^{p,u}(x^q) = d_T(x_u^p,x^q) / \sqrt{\bar{d_T^u}};~s_F^{p,u}(x^q) = d_F(x_u^p,x^q) / \sqrt{\bar{d_T^u}},
\end{equation}
\noindent where $p \in [1,n]_{\mathbb{Z}}$. After acquiring all scores, we compute the mean and minima of temporal scores $s_T^{u_{avg}}$, $s_T^{u_{min}}$ and frequency scores $s_F^{u_{avg}}$, $s_F^{u_{min}}$. We then use the frequency scores to adaptively weight the temporal scores, determining whether to accept the query:
\begin{equation}
	\label{Eq::threshold}
	s_T^{u_{min}}(1+sigmoid(s_F^{u_{min}})) + s_T^{u_{avg}}(1-sigmoid(s_F^{u_{avg}})) < c,
\end{equation}
\noindent where $c$ is a pre-defined threshold. If the distance summation fulfills Eq.~\ref{Eq::threshold}, the query $x^q$ is accepted as a genuine handwriting of writer $u$, otherwise it is determined as a forgery and rejected. By varying the threshold $c$, we can compute the Equal Error Rate metric (Sec.~\ref{sec::protocol}) for performance evaluation.

By harmonizing both temporal and frequency representations, MDV naturally fits in the multi-domain framework of SPECTRUM and amplifies the distinction between genuine samples and forgeries. This approach overcomes the limitation of relying solely on temporal features for verification in previous studies. Eq.~\ref{Eq::threshold} implies enhancing the more discriminative temporal scores while minimizing the less influential ones by adaptively re-weighting temporal scores with frequency scores. Importantly, these weights are dynamically derived from frequency features rather than being manually set, ensuring flexible adaptation to diverse handwriting scenarios.

\subsection{Model Optimization}
As described in Sec.~\ref{sec::mdv}, SPECTRUM performs verification using temporal and frequency feature representations. To optimize these representations, we employ a metric learning loss and a binary cross entropy loss with two key objectives: maximizing inter-class separation and minimizing intra-class variation.

As shown in Fig.~\ref{Fig::overall}, we first obtain temporal features $f_T \in \mathbb{R}^{L_T\times 64}$ by processing the input through two M$^3$I blocks, followed by a GRU and a Head module (a multi-layer perceptron). The features are then fed into a lifted-structure triplet loss \cite{triplet2016song} to separate genuine and forged samples in the embedding space. Assume that a batch of data contains handwriting from $N_w$ writers. For writer $u$ with $N_g$ genuine handwriting and $N_f$ forged handwriting, we randomly select one genuine handwriting of each writer as the anchor $x^u_a$, using the remaining $N_g - 1$ genuine handwriting as positives $x^u_{g,i}, i \in [1,...,N_g - 1]$ and $N_f$ forged handwriting as negatives $x^u_{f,j}, j \in [1,...,N_f]$. We construct the triplet pairs in the format of $(x^u_a,x^u_{g,i},x^u_{f,j})$, resulting in $(N_g - 1)\times N_f$ triplet pairs per writer. The loss for each triplet is defined as:
\begin{equation}
	l^u_{i,j}=max(0, d(x^u_a,x^u_{g,i}) + \epsilon - d(x^u_a,x^u_{f,j})),
\end{equation}
where $\epsilon$ is the positive margin. $d$ is the inner distance function, in which we use soft-DTW ($\gamma=5$) as the implementation following DsDTW \cite{dsdtw2022jiang}.  The triplet loss $L_{tri}$ is computed as:
\begin{equation}
	\mathcal{L}_{tri}^u = \frac{\sum^{N_g}_{i=1}\sum^{N_f}_{j=1}l^u_{i,j}}{|\{(i,j):l^u_{i,j}> 0\}| + 1};~\mathcal{L}_{tri} = \frac{1}{N_w}\sum^{N_w}_{u=1} \mathcal{L}_{tri}^u,
\end{equation}
where $|\{(i,j):l^u_{i,j}> 0\}|$ indicates the number of non-zero loss terms. Here, $L_{tri}$ aims to maximize the separation between genuine and forged handwriting, achieving a more discriminative embedding distribution. To further refine the representation, we introduce a regularization term $L_{intra}$ that minimizes intra-writer variations:
\begin{equation}
	L_{intra}^u = \frac{1}{N_g}\sum^{N_g}_{i=1}d(x^u_a,x^u_{g,i});~L_{intra} = \frac{1}{N_w}\sum^{N_w}_{u=1}L_{intra}^u.
\end{equation}

In addition, as shown in Fig.~\ref{Fig::overall}, the frequency feature maps from the last M$^3$I block undergo a selective pooling (SP) layer \cite{sig2vec2022lai} to derive frequency features $f_F$, which are subsequently converted to binary logits through the Head module. The logits are supervised by a binary cross entropy loss $\mathcal{L}_{BCE}$ (genuine sample$\rightarrow$label 1; forged sample$\rightarrow$label 0), further enhancing the discrimination between genuine and forged samples. 

The full optimization objective is formulated as:
\begin{equation}
	\mathcal{L} = \lambda\mathcal{L}_{intra}+\mathcal{L}_{tri}+\mathcal{L}_{BCE},
\end{equation}
where the regularization term's contribution is controlled by a weight $\lambda$. By default, we set $\lambda$ to 0.01 in all experiments.

\begin{table*}[t]
	\renewcommand{\arraystretch}{1.1}
	\caption{Comparison of SPECTRUM and existing methods on MSDS-ChS \cite{msds2022zhang}. Trans. denotes the Transformer architecture \cite{attention2017vaswani}. The best results are marked in \textbf{bold} and the second-best results are marked with \underline{underline}. The same hereafter.}
	\label{Table::chs}
	\centering
	\resizebox{0.9\textwidth}{!}{
		\begin{tabular}{l c c c c c c c c}
			\whline{1.05pt}
			\noalign{\vspace{3pt}}
			\multirow{2.4}{*}{Method} &
			\multirow{2.4}{*}{Venue} & \multirow{2.4}{*}{Architecture} & \multicolumn{4}{c}{Skilled Forgery $\downarrow$} & \multicolumn{2}{c}{Random Forgery $\downarrow$}\\
			\cmidrule(r){4-7}\cmidrule{8-9}
			~ & ~ & ~ & 4 vs 1 & 3 vs 1 & 2 vs 1 & 1 vs 1 & 4 vs 1 & 1 vs 1\\
			\hline
			\noalign{\vspace{2pt}}
			DTW \cite{vintsyuk1968speech} & - & - & 11.66/7.70 & 11.37/7.44 & 12.42/7.26 & 17.26/8.93 & \textbf{0.58}/0.20 & \textbf{1.03}/\underline{0.27}\\
			DeepDTW \cite{wu2019DeepDTW} & ICDAR'19 & CNN & 7.14/3.70 & 7.16/3.71 & 7.53/3.71 & 12.60/4.77 & 0.61/\underline{0.16} & 5.41/1.10\\
			TA-RNNs \cite{deepsign2021tbiom} & TBIOM'21 & RNN &  7.69/5.22 & 7.91/5.67 & 8.34/6.36 & 9.04/5.05 & 2.67/0.47 & \underline{1.55}/0.57\\
			Sig2Vec \cite{sig2vec2022lai} & TPAMI'22 & CNN & 9.03/4.97 & 8.78/4.92 & 9.87/5.16 & 15.10/7.27 & 1.93/0.74 & 5.09/1.18\\
			DsDTW \cite{dsdtw2022jiang} & TIFS'22 & CNN\&RNN & \underline{5.91/2.90} & \underline{5.69/2.90} & \underline{5.96/2.77} & \textbf{9.58/3.99} & 0.84/\textbf{0.11} & 1.87/\textbf{0.17}\\
			FBN \cite{fbn2023xie} & PR'23 & Trans. & 20.89/17.53 & 20.83/18.11 & 21.90/18.09 & 26.94/23.78 & 2.45/1.01 & 4.52/2.25\\
			ConvMixer \cite{convmixer2023fathy} & NILES'23 & CNN & 11.46/6.75 & 11.45/6.54 & 11.93/6.58 & 18.71/9.47 & 5.04/1.88 & 12.28/2.77\\
			MMHSV \cite{mmhsv2024icassp} & ICASSP'24 & CNN & 14.91/10.86 & 14.46/10.92 & 15.27/11.62 & 20.85/16.31 & 2.01/1.12 & 4.02/1.92\\
			HTCSigNet \cite{htcsignet2025pr} & PR'25 & CNN\&Trans. & 15.06/11.76 & 14.69/11.54 & 15.95/12.03 & 19.75/15.78 & 6.46/4.83 & 8.63/6.61\\
			\hline
			\rowcolor{lightred}
			SPECTRUM (\textbf{Ours}) & This work & CNN\&RNN & \textbf{5.30/2.47} & \textbf{5.33/2.53} & \textbf{5.88/2.62} & \underline{10.70/4.97} & \underline{0.72}/\textbf{0.11} & 2.72/0.32\\
			\whline{1.05pt}
	\end{tabular}}
\end{table*}

\begin{table*}[t]
	\renewcommand{\arraystretch}{1.1}
	\caption{Comparison of SPECTRUM and existing methods on MSDS-TDS \cite{msds2022zhang}.}
	\label{Table::tds}
	\centering
	\resizebox{0.9\textwidth}{!}{
		\begin{tabular}{l c c c c c c c c}
			\whline{1.05pt}
			\noalign{\vspace{2pt}}
			\multirow{2.4}{*}{Method} & \multirow{2.4}{*}{Venue} & \multirow{2.4}{*}{Architecture} & \multicolumn{4}{c}{Skilled Forgery $\downarrow$} & \multicolumn{2}{c}{Random Forgery $\downarrow$}\\
			\cmidrule(r){4-7}\cmidrule{8-9}
			~ & ~ & ~ & 4 vs 1 & 3 vs 1 & 2 vs 1 & 1 vs 1 & 4 vs 1 & 1 vs 1\\
			\hline
			\noalign{\vspace{3pt}}
			DTW \cite{vintsyuk1968speech} & - & - & 9.99/5.75 & 9.94/5.78 & 10.01/5.95 & 14.46/6.76 & \textbf{0.25/0.01} & \textbf{0.30}/\underline{0.04}\\
			DeepDTW \cite{wu2019DeepDTW} & ICDAR'19 & CNN & 5.75/1.94 & 5.60/1.93 & 5.49/1.95 & 9.56/2.11 & 0.63/0.28 & 5.16/0.40\\
			TA-RNNs \cite{deepsign2021tbiom} & TBIOM'21 & RNN &  5.11/2.91 & 5.44/3.06 & 5.77/3.16 & 5.94/2.60 & 1.71/0.40 & 0.85/0.21\\
			Sig2Vec \cite{sig2vec2022lai} & TPAMI'22 & CNN & 5.18/2.07 & 5.24/2.22 & 5.94/2.17 & 7.01/3.26 & 1.66/0.26 & 1.76/0.28\\
			DsDTW \cite{dsdtw2022jiang} & TIFS'22 & CNN\&RNN &  \underline{4.13/1.42} & \underline{4.05/1.41} & \underline{4.40/1.32} & \underline{5.76}/\textbf{1.85} & 0.42/0.07 & \underline{0.59}/0.14\\
			FBN \cite{fbn2023xie} & PR'23 & Trans. & 18.58/14.63 & 18.70/14.41 & 19.77/14.35 & 20.21/17.19 & 3.83/1.62 & 4.42/2.25\\
			ConvMixer \cite{convmixer2023fathy} & NILES'23 & CNN & 8.90/3.89 & 8.94/4.03 & 9.62/3.97 & 15.67/6.11 & 3.09/0.55 & 6.85/0.87\\
			MMHSV \cite{mmhsv2024icassp} & ICASSP'24 & CNN & 13.58/8.71 & 13.62/8.73 & 14.50/9.12 & 16.87/11.05 & 0.71/0.05 & 1.87/0.10\\
			HTCSigNet \cite{htcsignet2025pr} & PR'25 & CNN\&Trans. & 16.64/13.01 & 17.59/13.70 & 19.76/15.25 & 23.29/19.65 & 5.53/3.83 & 8.06/6.53\\
			\hline
			\rowcolor{lightred}
			SPECTRUM (\textbf{Ours}) & This work & CNN\&RNN & \textbf{3.38/1.20} & \textbf{3.48/1.11} & \textbf{3.57/1.18} & \textbf{5.20}/\underline{2.10} & \underline{0.30/0.04} & 0.76/\textbf{0.02}\\
			\whline{1.05pt}
	\end{tabular}}
\end{table*}

\section{Experiment}
\label{sec::experiment}
\subsection{Experiment Protocol}
\label{sec::protocol}
\paragraph{Dataset} We evaluate our method on three OHV datasets: MSDS-ChS (Chinese signatures) \cite{msds2022zhang}, MSDS-TDS (Token Digit Strings) \cite{msds2022zhang}, and DeepSignDB (Latin signatures) \cite{deepsign2021tbiom}. These are currently the largest public datasets for their respective handwriting types. Following an open-set setting, we ensure no overlap between training and testing users. MSDS-ChS and MSDS-TDS share 402 writers, which we split into 202 training and 200 testing users as per \cite{msds2022zhang}, resulting in 8,080/8,000 training/testing samples each. The two-session (across-session) data of each dataset is used by default. DeepSignDB consists of five subsets, in which, however, the Biosecure DS2 subset \cite{bsds22009tpami} currently releases only training data. We follow \cite{sig2vec2022lai,dsdtw2022jiang} and utilize the same subsets during training and testing, where the official ``development'' and ``evaluation'' sets of all subsets are utilized as training/testing data, respectively. This results in 21,104/20,596 training/testing samples from 528/512 users.

\vspace{-8pt}
\paragraph{Metric} We adopt Equal Error Rate (EER) as the evaluation metric, the point at which False Acceptance Rate equals False Rejection Rate. The proposed MDV is employed to compute EER\%, with details provided in Sec.~\ref{sec::mdv}. Following the protocols of MSDS and DeepSignDB, we report EERs under both global and local (user-specific) thresholds, displaying the results as $EER_g/EER_l$ on MSDS-ChS and MSDS-TDS. For DeepSignDB, we report only EERs under the global threshold. All results are reported in percentage.

\vspace{-5pt}
\paragraph{Impostor types} We consider both skilled and random forgeries as impostor types. Skilled forgeries are drawn from the skillfully forged samples provided in the datasets, while random forgeries consist of the genuine samples of other writers.

\vspace{-5pt}
\paragraph{Template selection} The number of genuine templates significantly impacts verification performance. We follow the protocols of MSDS \cite{msds2022zhang} and DeepSignDB \cite{deepsign2021tbiom} to select templates. For MSDS-ChS and MSDS-TDS, we use one to four templates against a single query in skilled forgery verification (4 vs 1, 3 vs 1, 2 vs 1, and 1 vs 1), and four and one templates in random forgery verification. For DeepSignDB, we employ four or one templates for both skilled and random forgery scenarios. To ensure reproducibility, we consistently select the first $n$ genuine samples as templates.


More details regarding data preprocessing and implementation are included in Appendix~\ref{appendix::preprocess} and Appendix~\ref{appendix::impl}, respectively.

\begin{table*}[t]
	\renewcommand{\arraystretch}{1.0}
	\caption{Comparison of SPECTRUM and existing methods on DeepSignDB \cite{deepsign2021tbiom}.}
	\label{Table::deepsign}
	\centering
	\resizebox{0.85\textwidth}{!}{
		\begin{tabular}{l c c c c c c c c c c}
			\whline{1.05pt}
			\noalign{\vspace{2pt}}
			\multirow{3.6}{*}{Method} & \multirow{3.6}{*}{Venue} & \multirow{3.6}{*}{Architecture} & \multicolumn{4}{c}{Stylus} & \multicolumn{4}{c}{Finger}\\
			\cmidrule(r){4-7}\cmidrule{8-11}
			~ & ~ & ~ & \multicolumn{2}{c}{Skilled Forgery $\downarrow$} & \multicolumn{2}{c}{Random Forgery $\downarrow$} & \multicolumn{2}{c}{Skilled Forgery $\downarrow$} & \multicolumn{2}{c}{Random Forgery $\downarrow$}\\
			\cmidrule(r){4-5}\cmidrule(r){6-7}\cmidrule(r){8-9}\cmidrule(r){10-11}
			~ & ~ & ~ & 4 vs 1 & 1 vs 1 & 4 vs 1 & 1 vs 1 & 4 vs 1 & 1 vs 1 & 4 vs 1 & 1 vs 1\\
			\hline
			\noalign{\vspace{3pt}}
			DTW \cite{vintsyuk1968speech} & - & - & 4.53 & 7.06 & 1.23 & 1.98 & 10.66 & 14.74 & 1.02 & 1.25\\
			DeepDTW \cite{wu2019DeepDTW} & ICDAR'19 & CNN & 2.97 & 5.98 & 1.63 & 3.13 & 7.02 & 12.27 & 2.78 & 5.17\\
			TA-RNNs \cite{deepsign2021tbiom} & TBIOM'21 & RNN & 3.30 & 4.20 & \underline{0.60} & \underline{1.50} & 11.30 & 13.80 & \underline{1.00} & \underline{1.80}\\
			Sig2Vec \cite{sig2vec2022lai} & TPAMI'22 & CNN & \textbf{2.54} & \underline{4.08} & \textbf{0.48} & \textbf{0.84} & \underline{6.97} & \textbf{10.87} & \textbf{0.79} & \textbf{1.86}\\
			DsDTW \cite{dsdtw2022jiang} & TIFS'22 & CNN\&RNN & \textbf{2.54} & \textbf{4.04} & 0.97 & 1.69 & 6.99 & 11.84 & 1.81 & 2.89\\
			FBN \cite{fbn2023xie} & PR'23 & Trans. & 13.60 & 15.41 & 2.25 & 3.01 & 20.82 & 23.11 & 3.43 & 5.26\\
			ConvMixer \cite{convmixer2023fathy} & NLES'23 & CNN & 8.08 & 17.03 & 6.21 & 11.67 & 13.85 & 20.24 & 7.03 & 11.22\\
			MMHSV \cite{mmhsv2024icassp} & ICASSP'24 & CNN & 11.38 & 17.43 & 4.34 & 8.62 & 16.27 & 21.03 & 5.71 & 8.65\\
			HTCSigNet \cite{htcsignet2025pr} & PR'25 & CNN\&Trans. & 9.53 & 12.75 & 7.15 & 9.98 & 19.13 & 23.20 & 11.25 & 14.85\\
			\hline
			\rowcolor{lightred}
			SPECTRUM (\textbf{Ours}) & This work & CNN\&RNN & \underline{2.61} & 4.31 & 1.13 & 1.99 & \textbf{6.96} & \underline{11.44} & 2.38 & 4.63\\
			\whline{1.05pt}
	\end{tabular}}
\end{table*}

\begin{table*}[t]
	\renewcommand{\arraystretch}{1.05}
	\caption{Ablation study on MSDS-TDS \cite{msds2022zhang} and MSDS-ChS \cite{msds2022zhang}. \emph{Baseline} indicates a model consists of merely two Conv modules (Fig.~\ref{Fig::overall}) and a GRU. \emph{Frequency} denotes introducing a single-scale interactor for frequency modeling. $\times$ indicates the removal of specific modules, except for replacing the self-gated fusion module with the addition operation.}
	\label{Table::ablation}
	\centering
	\resizebox{\textwidth}{!}{
		\begin{tabular}{c c c c c c c c c c | c c c c}
			\whline{1.05pt}
			\noalign{\vspace{2pt}}
			\multirow{3.8}{*}{\#Line} &
			\multirow{3.8}{*}{Baseline} & \multirow{3.8}{*}{Frequency} & \multirow{3.8}{*}{Multi-Scale} &
			\multirow{3.8}{*}{Self-Gated Fusion} & \multirow{3.8}{*}{MDV} &
			\multicolumn{4}{c|}{MSDS-TDS} & \multicolumn{4}{c}{MSDS-ChS}\\
			\cmidrule(r){7-10}\cmidrule(r){11-14}
			~ & ~ & ~ & ~ & ~ & ~ & \multicolumn{2}{c}{Skilled Forgery $\downarrow$} & \multicolumn{2}{c|}{Random Forgery $\downarrow$} & \multicolumn{2}{c}{Skilled Forgery $\downarrow$} & \multicolumn{2}{c}{Random Forgery $\downarrow$}\\
			\cmidrule(r){7-8}\cmidrule(r){9-10}\cmidrule(r){11-12}\cmidrule(r){13-14}
			~ & ~ & ~ & ~ & ~ & ~ & 4 vs 1 & 1 vs 1 & 4 vs 1 & 1 vs 1 & 4 vs 1 & 1 vs 1 & 4 vs 1 & 1 vs 1\\
			\hline
			\noalign{\vspace{1pt}}
			1 & \checkmark & $\times$ & $\times$ & $\times$ & $\times$ & 4.13/1.30 & 6.09/2.09 & 0.36/0.05 & 1.21/0.08 & 5.98/2.80 & 11.30/5.13 & 1.19/0.22 & 4.25/0.57\\
			\hline
			2 & \checkmark & \checkmark & $\times$ & $\times$ & $\times$ & 5.02/1.38 & 7.28/2.39 & 0.49/0.08 & 1.39/0.09 & 6.50/2.91 & 11.22/4.94
			& 0.98/0.14 & 3.35/0.36\\
			\hline
			3 & \checkmark & \checkmark & $\times$ & $\times$ & \checkmark & 4.95/1.36 & 7.28/2.39 & 0.50/0.09 & 1.39/0.09 & 6.13/2.86 & 11.22/4.94 & 0.93/0.15 & 3.35/0.36\\
			\hline
			4 & \checkmark & \checkmark & \checkmark & $\times$ & \checkmark & 4.05/1.43 & 5.90/2.07 & 0.34/0.04 & \textbf{0.70}/0.03 & 5.49/2.45 & 10.40/4.68 & 0.90/0.17 & 3.22/0.47\\
			\hline
			5 & \checkmark & \checkmark & $\times$ & \checkmark & \checkmark & 4.67/1.46 & 7.02/2.25 & 0.59/0.05 & 1.54/0.08 & 6.20/3.12 & 12.33/5.85 & 1.05/0.14 & 3.96/0.54\\
			\hline
			6 & \checkmark & \checkmark & \checkmark & \checkmark & $\times$ & 3.44/1.22 & 5.20/2.10 & \textbf{0.25/0.04} & 0.76/\textbf{0.02} & 5.51/2.75 & 10.70/4.97 & 0.74/\textbf{0.10} & \textbf{2.72/0.32}\\
			\hline
			7 & \checkmark & \checkmark & \checkmark & \checkmark & \checkmark & \textbf{3.38/1.20} & \textbf{5.20/2.10} & 0.30/\textbf{0.04} & 0.76/\textbf{0.02} & \textbf{5.30/2.47} & \textbf{10.70/4.97} & \textbf{0.72}/0.11 & \textbf{2.72/0.32}\\
			\whline{1.05pt}
	\end{tabular}}
\end{table*}

\subsection{Comparison with State-of-the-Art Method}
We evaluate SPECTRUM's OHV performance against existing methods on MSDS-ChS, MSDS-TDS, and DeepSignDB in Tables~\ref{Table::chs} to \ref{Table::deepsign}. Baselines include: (1) DTW \cite{vintsyuk1968speech}, a non-trained Dynamic Time Warping approach; (2) state-of-the-art (SOTA) online signature verification models DeepDTW \cite{wu2019DeepDTW}, TA-RNNs \cite{deepsign2021tbiom}, Sig2Vec \cite{sig2vec2022lai}, DsDTW \cite{dsdtw2022jiang}, and ConvMixer \cite{convmixer2023fathy}; (3) MMHSV \cite{mmhsv2024icassp}, adapted for online handwriting by replacing the audio input; and (4) Transformer-based models FBN \cite{fbn2023xie} (BERT-based) and HTCSigNet \cite{htcsignet2025pr} (a hybrid CNN-Transformer migrated from offline verification). From the results, we can draw the following observations.

(1) As evidenced in Tables~\ref{Table::chs} and \ref{Table::tds}, SPECTRUM generally outperforms existing methods on MSDS-ChS and MSDS-TDS. Under skilled forgery scenarios, it achieves 5.30/2.47 ($EER_g/EER_l$) on MSDS-ChS and 3.38/1.20 on MSDS-TDS, significantly exceeding the second-best performance of 5.91/2.90 and 4.13/1.42. Under random forgery scenarios, SPECTRUM outstrips other methods like DsDTW and Sig2Vec on both datasets, especially on MSDS-TDS. Although the DTW method slightly edges out SPECTRUM, the margin is narrow and does not diminish SPECTRUM's competitiveness. The outperformance is primarily attributed to SPECTRUM's dual-domain learning approach, which integrates temporal and frequency features, resulting in a more robust handwriting representation compared to single-domain methods.

(2) Table~\ref{Table::deepsign} demonstrates that SPECTRUM delivers comparable performance compared to SOTA methods on DeepSignDB. Although the Sig2Vec model primarily holds sway, our SPECTRUM exhibits the best/second-best results in some cases, such as in the skilled forgery verification based on stylus-/finger-written signatures. Compared to performance on MSDS-ChS and MSDS-TDS, the relatively lower results on DeepSignDB could be attributed to two aspects. 1) \textbf{Length variations.} DeepSignDB exhibits substantially larger length variations (range: $45\sim311,819$, $\sigma=578.65$) compared to MSDS-ChS (range: $208\sim34,294$, $\sigma=427.56$) and MSDS-TDS (range: $300\sim3,504$, $\sigma=230.10$). The pronounced length variations likely introduce additional verification challenges. 2) \textbf{Stroke discrepancy.} Latin signatures in DeepSignDB typically comprise continuous, scribble-like strokes, unlike the discrete strokes in Chinese and TDS signatures. SPECTRUM's frequency modeling could be more effective for discrete strokes than continuous ones, potentially contributing to the model's inferior performance.

(3) As observed, CNN-/RNN-based models dominate the SOTA performance, while Transformer-based models significantly lag behind. We speculate that this disparity stems from two key factors. 1) \textbf{Attention inefficiency.} The self-attention mechanism of Transformer is adept at capturing global sequence dependency, while may not be efficient in learning local writing patterns crucial for OHV. CNN/RNN architectures could better capture these local fine-grained features. 2) \textbf{Limited data volume.} The training data volume is 8,080 for each MSDS subset and 21,104 for DeepSignDB, which may be insufficient to meet the large data requirements of Transformer models and limit their generalization abilities. Notably, by combining Transformer with CNN, HTCSigNet outperforms the pure Transformer-based FBN, validating the better adaptiveness of the CNN architecture for OHV.

(4) Most methods, including our SPECTRUM, perform better on MSDS-TDS than on MSDS-ChS. Note that MSDS-ChS and MSDS-TDS are collected from the same 402 users and share identical data splitting, ensuring fair comparisons. This resonates with the phenomenon discovered in \cite{msds2022zhang} that the accuracy of TDS verification is higher than that of Chinese signature verification, reinforcing that TDS could be a more effective and reliable handwritten biometric than Chinese signature.

Additionally, we perform comparisons on model efficiency including inference speed and model parameters in Appendix~\ref{appendix::efficiency}. We further demonstrate SPECTRUM's effectiveness via feature visualizations in Appendix~\ref{appendix::visualization}.

\begin{table*}[t]
	\renewcommand{\arraystretch}{1.0}
	\caption{Multi-biometric fusion of Chinese signatures from MSDS-ChS \cite{msds2022zhang} and Token Digit Strings from MSDS-TDS \cite{msds2022zhang}.}
	\label{Table::biometric}
	\centering
	\resizebox{0.8\linewidth}{!}{
		\begin{tabular}{c c c c c c c c c}
			\whline{1.05pt}
			\noalign{\vspace{2pt}}
			\multirow{2.4}{*}{Method} & \multirow{2.4}{*}{Venue} & \multirow{2.4}{*}{Biometric} & \multicolumn{4}{c}{Skilled Forgery $\downarrow$} & \multicolumn{2}{c}{Random Forgery $\downarrow$}\\
			\cmidrule(r){4-7}\cmidrule{8-9}
			~ & ~ & ~ & 4 vs 1 & 3 vs 1 & 2 vs 1 & 1 vs 1 & 4 vs 1 & 1 vs 1\\
			\hline
			\noalign{\vspace{3pt}}
			\multirow{3.8}{*}{Sig2Vec \cite{sig2vec2022lai}} & \multirow{3.8}{*}{TPAMI'22} & ChS & 9.03/4.97 & 8.78/4.92 & 9.87/5.16 & 15.10/7.27 & 1.93/0.74 & 5.09/1.18\\
			\cmidrule(r){3-9}
			~ & ~ & TDS & 5.18/2.07 & 5.24/2.22 & 5.94/2.17 & 7.01/3.26 & 1.66/0.26 & 1.76/0.28\\
			\cmidrule(r){3-9}
			~ & ~ & Both & 5.04/1.83 & 5.23/1.83 & 5.28/1.78 & 8.89/2.96 & 0.63/0.12 & 1.42/0.20\\
			
			\cmidrule(r){1-9}
			\multirow{3.8}{*}{DsDTW \cite{dsdtw2022jiang}} & \multirow{3.8}{*}{TIFS'22} & ChS & 5.91/2.90 & 5.69/2.90 & 5.96/2.77 & 9.58/3.99 & 0.84/0.11 & 1.87/0.17\\
			\cmidrule(r){3-9}
			~ & ~ & TDS & 4.13/1.42 & 4.05/1.41 & 4.40/1.32 & 5.76/1.85 & 0.42/0.07 & \textbf{0.59}/0.14\\
			\cmidrule(r){3-9}
			~ & ~ & Both & 3.77/0.89 & 3.65/0.93 & 3.80/1.03 & 6.22/2.08 & \textbf{0.15/0.03} & 0.94/0.16\\
			\cmidrule(r){1-9}
			
			\multirow{3.8}{*}{SPECTRUM (\textbf{Ours})} & \multirow{3.8}{*}{This work} & ChS & 5.30/2.47 & 5.33/2.53 & 5.88/2.62 & 10.70/4.97 & 0.72/0.11 & 2.72/0.32\\
			\cmidrule(r){3-9}
			~ & ~ & TDS & 3.38/1.20 & 3.48/1.11 & 3.57/1.18 & 5.20/2.10 & 0.30/\underline{0.04} & \underline{0.76}/\textbf{0.02}\\
			\cmidrule(r){3-9}
			\rowcolor{lightred}
			\noalign{\vspace{-1pt}}
			~ & ~ & Both & \textbf{3.15/0.80} & \textbf{3.11/0.81} & \textbf{3.23/0.78} & \textbf{5.76/1.25} & \underline{0.21}/0.05 & 1.08/\underline{0.06}\\
			\whline{1.05pt}
	\end{tabular}}
\end{table*}

\subsection{Ablation Study}
\label{sec::ablation}
We conduct ablation studies to evaluate the effectiveness of different components inside SPECTRUM on MSDS-TDS and MSDS-ChS. \emph{Baseline} consists of merely two \emph{Conv} modules (Fig.~\ref{Fig::overall}) and a GRU. \emph{Frequency} refers to incorporating a single-scale interactor rather than a multi-scale interactor. $\times$ indicates the removal of specific modules, except where the self-gated fusion module is replaced by addition. Results are summarized in Table~\ref{Table::ablation}.

Comparing Lines 1 and 2, we observe that a single-scale interactor initially impairs model performance. However, Lines 3 and 4 reveal that using the multi-scale interactor rather than the single-scale one significantly improves model performance, evidenced by the gains of 0.90\%/0.64\% (global threshold; skilled forgery; the same hereafter) in the most difficult skilled forgery scenario on MSDS-TDS and MSDS-ChS, respectively. Furthermore, comparing Lines 5 and 7, removing the multi-scale interactor results in 1.29\% and 0.90\% declines. These outcomes strongly demonstrate the significance of the multi-scale interactor in introducing fine-grained frequency features and enhancing handwriting representations. In addition, the self-gated fusion module brings 0.67\% and 0.19\% improvements on the two datasets, respectively (Lines 4 and 7). The MDV further boosts performance by 0.06\% and 0.21\% (Lines 6 and 7). Ultimately, incorporating all designs yields the best performance, confirming the effectiveness of SPECTRUM's modules and the benefits of our multi-domain learning approach.

We conduct more ablation studies on the number of scales $m$ of the multi-scale interactor and different gated mechanisms within the self-gated module, which are included in Appendix~\ref{appendix::extended_abl}. Additionally, we investigate the decision-making behavior of the self-gated fusion module, with results and discussion presented in Appendix~\ref{appendix::selfgated}.

\subsection{Biometric-Based Multi-Domain Representation Learning}
We further investigate multi-domain learning from the perspective of multiple biometric mediums. Since the Chinese signature (ChS) in MSDS-ChS \cite{msds2022zhang} and Token Digit String (TDS) \cite{msds2022zhang} in MSDS-TDS come from the same writers, it offers a natural avenue to incorporate both ChS and TDS to explore their collaborative potential for OHV. Therefore, we design a dual-path architecture where each path processes either Chinese signatures (ChS) or Token Digit Strings (TDS) using identical model structures. Two well-performed OHV models (Sig2Vec \cite{sig2vec2022lai}, DsDTW \cite{dsdtw2022jiang}) and the proposed SPECTRUM are applied in this dual-path architecture for experiments. The sequence representations are concatenated along spatial dimensions and the logits are averaged from the two paths for optimization and inference. Following the split in Sec.~\ref{sec::protocol}, the data of MSDS-ChS and MSDS-TDS is merged to create consolidated training and testing sets while maintaining the open-set setting. Experimental results are presented in Table~\ref{Table::biometric}. 

As observed, combining ChS and TDS generally improves performance over using either biometric alone across all three methods, particularly in the most challenging skilled forgery scenario. The improvement brings forth several key insights. (1) Combining multiple handwritten biometrics improves verification performance. This is likely due to the richer, more expressive feature space, which amplifies individual stylistic representations and enhances models' discriminatory power. (2) Under the combined-biometric context, SPECTRUM attains consistently optimal results in skilled forgery verification and near-top results in random forgery verification. This demonstrates SPECTRUM's ability to perform multi-domain learning across both feature domains (temporal and frequency) and biometric domains, bolstering performance through the unprecedented synergy of multiple features and biometrics. (3) Even basic biometric fusion strategies like concatenation can yield performance improvements. This suggests significant potential for developing more sophisticated feature fusion mechanisms to better leverage the commonalities between different handwritten biometrics, pointing out a promising direction for future research.

\section{Conclusion}
In this paper, we propose SPECTRUM, a novel OHV model driven by multi-domain representation learning. We propose a multi-scale interactor for blending local temporal and frequency features across multiple spatial scales, coupled with a self-gated fusion module that integrates global temporal-frequency features through self-balancing. In addition, we design a multi-domain distance-based verifier that naturally harnesses both temporal and frequency representations to sharpen the distinction between genuine and forged samples. Extensive experiments demonstrate the superior performance of SPECTRUM over existing OHV methods. Additionally, we discover that combining multiple handwritten biometrics fundamentally improves feature discrimination. These findings not only validate the effectiveness of multi-domain representation learning across both feature and biometric domains but also suggest promising new directions for future research to enhance the reliability and real-world applicability of OHV systems. Limitations and further discussions of this work are included in Appendix~\ref{appendix::limitation}.

\begin{acks}
This research is supported in part by the National Key Research and Development Program of China (2022YFC3301702, 2022YFC3301703), and the National Natural Science Foundation of China (Grant No.: 62476093).
\end{acks}

\bibliographystyle{ACM-Reference-Format}
\balance
\bibliography{reference}

\clearpage
\appendix

\section*{Appendix}

\section{Data Preporcessing}
\label{appendix::preprocess}
\begin{table}[h]
	\renewcommand{\arraystretch}{1.1}
	\caption{Time-function features.}
	\label{Table::timefunc}
	\centering
	\resizebox{0.9\linewidth}{!}{
		\begin{tabular}{c c}
			\whline{1.1pt}
			\noalign{\vspace{1pt}}
			\textbf{\#} & \textbf{Features}\\
			\hline
			1-2 & Horizontal and vertical component velocity $x$, $y$: $\dot{x},\dot{y}$\\
			3-4 & Line velocity and acceleration: $v = \sqrt{\dot{x}^2 + \dot{y}^2},\dot{v}$\\
			5 & Path-tangent angle: $\theta = \arctan{\frac{\dot{y}}{\dot{x}}}$\\
			6-7 & Cosine and sine of angle: $\cos{\theta},\sin{\theta}$\\
			8-9 & Angular velocity and acceleration:  $\dot{\theta},\ddot{\theta}$\\
			11 & Centripetal acceleration magnitude: $\bigtriangleup v =  v \cdot \dot{\theta}$\\
			12 & Total acceleration magnitude: $a = \sqrt{\dot{v}^2 + \bigtriangleup v^2}$\\
			13-15 & Pressure and its first- and second-order derivatives: $p,\dot{p},\ddot{p}$\\
			\whline{1.1pt}
	\end{tabular}}
\end{table}
We utilize the $x$, $y$ coordinates, and pressure $p$ of the raw online handwritten data for preprocessing. To mitigate variations in size and location, we apply center normalization on $x$ and $y$, relocating the writing center to (0,0) and scaling coordinates to (–1, 1) while preserving aspect ratio. Pressure values are normalized via min-max scaling. Subsequently, following the original settings in \cite{msds2022zhang,deepsign2021tbiom}, we resample the data in MSDS-ChS and MSDS-TDS into 120Hz and the data in DeepSignDB into 100Hz, using bi-cubic interpolation. We extract 15 time-function features based on the normalized $x$, $y$, and $p$ as model input, as outlined in Table~\ref{Table::timefunc}. All time-function features are standardized using z-score normalization to have zero mean and unit variance.

\section{Implementation Detail}
\label{appendix::impl}
We train SPECTRUM for 40 epochs using AdamW with $\beta_1 = 0.9$, $\beta_2 = 0.999$, and a weight decay of 1e-2. The learning rate starts at 5e-4 and decays to 5e-7 following a cosine schedule. Each batch randomly samples handwriting from four writers, including five genuine samples, five skilled forgeries, and five random forgeries per writer, yielding a batch size of $4 \times (5 + 5 \times 2) = 60$. Genuine and skilled forgeries are taken directly from the dataset, while random forgeries are genuine samples from five other writers. The pre-defined threshold $c$ in Eq.~\ref{Eq::threshold} is uniformly sampled from 0 to 50 with a step size of 0.01.

\section{Model Efficiency}
\label{appendix::efficiency}
We conduct a comparative analysis regarding inference speed and model parameters between SPECTRUM and existing models, as illustrated in Table~\ref{Table::comp_inference}. During verification, we compute sample-wise Euclidean distance for models that output one-dimensional feature vectors, while computing dynamic time warping (DTW) distance for those output two-dimensional temporal representations. All model inferences are conducted on a machine with one RTX 3090 24GB GPU and a 6-core Intel Core i5-8600K CPU.

From the results of inference speed (the penultimate column), it can be observed that CNN models using Euclidean distance, such as Sig2Vec, ConvMixer, and MMHSV, achieve the fastest speed. In contrast, models relying on DTW distance, including DeepDTW, DsDTW, and SPECTRUM, require significantly more computation time. Nevertheless, SPECTRUM achieves a higher inference speed of 6.97ms/s than DeepDTW (9.48ms/s) and DsDTW (13.57ms/s), substantiating its efficiency. In addition, model architecture significantly impacts computational efficiency. Even using the Euclidean distance, the RNN- and Transformer-based models, \emph{e.g.}, TA-RNNs and FBN, are significantly slower than models built with CNN. In terms of model size, SPECTRUM boasts a modest footprint of 1.36M parameters. While not the most compact model, SPECTRUM’s parameter count remains minimal. Collectively, both the fast inference speed and modest parameter size demonstrate the computational and storage efficiency of the proposed SPECTRUM.

\begin{table}[ht]
	\renewcommand{\arraystretch}{1.1}
	\caption{Comparison of the inference speed and model parameters between SPECTRUM and other methods. The inference is conducted on the test set of MSDS-TDS (8000 samples) \cite{msds2022zhang} under the 4 vs 1 skilled forgery scenario. The inference time includes feature extraction time and verification time. Distance represents the distance computed during verification, in which ``Euclidean'' denotes Euclidean distance, whereas ``DTW'' denotes the dynamic time warping distance. ``T.'' denotes Time. ``/s'' denotes per sample.}
	\label{Table::comp_inference}
	\centering
	\resizebox{\linewidth}{!}{
		\begin{tabular}{l c c c c c}
			\whline{1.05pt}
			Method & Venue & Architecture & Distance & T.\tiny{Inference} \normalsize (ms/s) & \#Params\\
			\hline
			DeepDTW \cite{wu2019DeepDTW} & ICDAR'19 & CNN & DTW & 9.48 & 30.53K\\
			TA-RNNs \cite{deepsign2021tbiom} & TBIOM'21 & RNN & Euclidean & 12.38 & 84.09K \\
			Sig2Vec \cite{sig2vec2022lai} & TPAMI'22 & CNN & Euclidean & 0.23 & 655.74K\\
			DsDTW \cite{dsdtw2022jiang} & TIFS'22 & CNN\&RNN & DTW & 13.57 & 236.48K\\
			FBN \cite{fbn2023xie} & PR'23 & Trans. & Euclidean & 13.98 & 43.27M\\
			ConvMixer \cite{convmixer2023fathy} & NILES'23 & CNN & Euclidean & 0.18 & 2.21M\\
			MMHSV \cite{mmhsv2024icassp} & ICASSP'24 & CNN & Euclidean & 0.77 & 1.64M\\
			HTCSigNet \cite{htcsignet2025pr} & PR'25 & CNN\&Trans. & Euclidean & 2.83 & 3.98M\\
			\hline
			\rowcolor{lightred}
			SPECTRUM (\textbf{Ours}) & This work & CNN\&RNN & DTW\&Euclidean & 6.97 & 1.36M\\
			\whline{1.05pt}
	\end{tabular}}
\end{table}

\section{Ablation Study on Module Implementation}
\label{appendix::extended_abl}
Except for evaluating different modules' effectiveness in Sec.~\ref{sec::ablation}, we further investigate the effect of different module implementations inside SPECTRUM. As described in Sec.~\ref{sec::m3i}, the number of scales used in the multi-scale interactor could potentially impact model performance. Therefore, we perform assessments by setting two, three, and four scales in the multi-scale interactor, with results presented in Table~\ref{Table::ablationmulti}. As observed, using three scales yields general optimal results on MSDS-TDS. While using four scales performs better on MSDS-ChS in the skilled forgery scenario, the three-scale configuration closely trails behind and yields optimal performance in the random forgery scenario. Hence, we adopt the three-scale configuration in SPECTRUM as the optimal general-purpose choice. 

Furthermore, we evaluate the impact of using sigmoid versus softmax as the gated mechanism within the self-gated module, as shown in Table~\ref{Table::ablationgated}. The results reveal that using softmax significantly degrades model performance compared to using sigmoid as the gated function. This could originate from sigmoid's smoother distribution, which better balances temporal and frequency features, enabling improved feature fusion and enhanced model performance.

\begin{figure*}[t]
	\centering
	\includegraphics[width=0.8\linewidth]{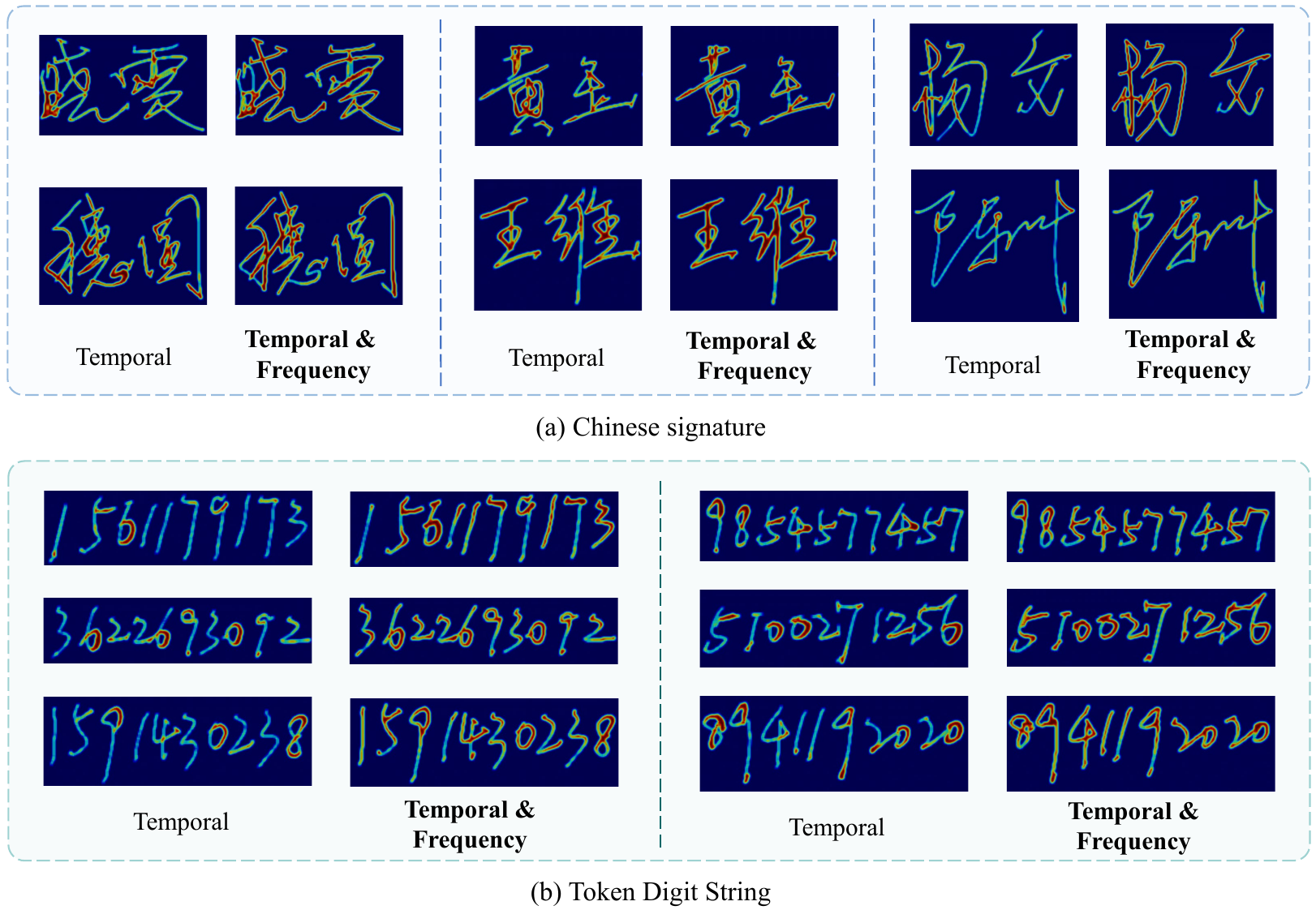}{}
	\caption{Visualization of the final feature representations on Chinese signature and Token Digit String data from MSDS-ChS and MSDS-TDS \cite{msds2022zhang}. The ``Temporal'' features are outputted by the Baseline model (as described in Sec.~\ref{sec::ablation}) that merely involves temporal domain learning, while the ``\textbf{Temporal \& Frequency}'' features are obtained from our SPECTRUM. The handwritten data are desensitized through cropping to protect privacy.}
	\label{Fig::visualization}
\end{figure*}

\begin{table}[t]
	\renewcommand{\arraystretch}{1.1}
	\caption{Ablation study regarding the number of scales in the multi-scale interactor.}
	\label{Table::ablationmulti}
	\centering
	\resizebox{\linewidth}{!}{
		\begin{tabular}{c c c c c | c c c c}
			\whline{1.05pt}
			\noalign{\vspace{1pt}}
			\multirow{3.5}{*}{\#Scales} &
			\multicolumn{4}{c|}{MSDS-TDS} & \multicolumn{4}{c}{MSDS-ChS}\\
			\noalign{\vspace{-1pt}}
			\cmidrule(r){2-5}\cmidrule(r){6-9}
			\noalign{\vspace{-1pt}}
			~ & \multicolumn{2}{c}{Skilled Forgery $\downarrow$} & \multicolumn{2}{c|}{Random Forgery $\downarrow$} & \multicolumn{2}{c}{Skilled Forgery $\downarrow$} & \multicolumn{2}{c}{Random Forgery $\downarrow$}\\
			\noalign{\vspace{-1pt}}
			\cmidrule(r){2-3}\cmidrule(r){4-5}\cmidrule(r){6-7}\cmidrule(r){8-9}
			\noalign{\vspace{-1pt}}
			~ & 4 vs 1 & 1 vs 1 & 4 vs 1 & 1 vs 1 & 4 vs 1 & 1 vs 1 & 4 vs 1 & 1 vs 1\\
			\noalign{\vspace{-1pt}}
			\hline
			\noalign{\vspace{1pt}}
			2 & 3.68/1.26 & 5.61/2.07 & 0.37/0.03 & 0.90/0.05 & 5.26/2.57 & 10.61/5.06 & 0.74/0.13 & 3.08/0.42\\
			\hline
			3 & \textbf{3.38}/1.20 & \textbf{5.20}/2.10 & \textbf{0.30/0.04} & \textbf{0.76/0.02} & 5.30/\textbf{2.47} & {10.70/4.97} & \textbf{0.72/0.11} & \textbf{2.72/0.32}\\
			\hline
			4 & 3.81/\textbf{1.12} & 6.09/\textit{1.96} & 0.38/0.06 & 1.14/0.06 & \textbf{5.20}/2.57 & \textbf{10.53/4.74} & 0.80/0.20 & 2.83/0.45\\
			\whline{1.05pt}
	\end{tabular}}
\end{table}

\begin{table}[t]
	\renewcommand{\arraystretch}{1.1}
	\caption{Ablation study regarding the gate implementation of the self-gated fusion module.}
	\label{Table::ablationgated}
	\centering
	\resizebox{\linewidth}{!}{
		\begin{tabular}{c c c c c | c c c c}
			\whline{1.05pt}
			\noalign{\vspace{1pt}}
			\multirow{3.5}{*}{Gate} &
			\multicolumn{4}{c|}{MSDS-TDS} & \multicolumn{4}{c}{MSDS-ChS}\\
			\noalign{\vspace{-1pt}}
			\cmidrule(r){2-5}\cmidrule(r){6-9}
			\noalign{\vspace{-1pt}}
			~ & \multicolumn{2}{c}{Skilled Forgery $\downarrow$} & \multicolumn{2}{c|}{Random Forgery $\downarrow$} & \multicolumn{2}{c}{Skilled Forgery $\downarrow$} & \multicolumn{2}{c}{Random Forgery $\downarrow$}\\
			\noalign{\vspace{-1pt}}
			\cmidrule(r){2-3}\cmidrule(r){4-5}\cmidrule(r){6-7}\cmidrule(r){8-9}
			\noalign{\vspace{-1pt}}
			~ & 4 vs 1 & 1 vs 1 & 4 vs 1 & 1 vs 1 & 4 vs 1 & 1 vs 1 & 4 vs 1 & 1 vs 1\\
			\noalign{\vspace{-1pt}}
			\hline
			\noalign{\vspace{1pt}}
			Softmax & 6.38/2.97 & 8.85/4.57 & 1.14/0.27 & 2.08/0.32 & 7.51/4.35 & 13.26/8.15 & 1.54/0.40 & 4.06/1.24\\
			\hline
			Sigmoid & \textbf{3.38/1.20} & \textbf{5.20/2.10} & \textbf{0.30/0.04} & \textbf{0.76/0.02} & \textbf{5.30/2.47} & \textbf{10.70/4.97} & \textbf{0.72/0.11} & \textbf{2.72/0.32}\\
			\whline{1.05pt}
	\end{tabular}}
\end{table}

\section{Feature Preference Analysis of the Self-Gated Fusion Module}
\label{appendix::selfgated}
The self-gated fusion module is designed to weight the importance of temporal and frequency features in a self-driven manner. To understand its decision-making process, we analyze the distributions of gate values across different datasets. Specifically, we compute the gate values on the testing data of each dataset, averaging them to obtain a scalar for comparisons against the neutral threshold of 0.5. As shown in Eq.~\ref{Eq::gate}, average above 0.5 indicates the preference for temporal features, while values below 0.5 represent the choice of frequency features.

The results reveal distinct preferences on different datasets:

$\bullet$ MSDS-ChS: 83\% temporal / 17\% frequency

$\bullet$ MSDS-TDS: 86\% temporal / 14\% frequency

$\bullet$ DeepSignDB: 41\% temporal / 59\% frequency

These results suggest that handwriting involving discrete, sharp strokes, such as Chinese signatures and digit strings (MSDS-ChS and MSDS-TDS), tends to emphasize temporal features. This emphasis could stem from the rich temporal clues present in stroke-level motion features like velocity and acceleration, as well as sequential dynamics like transitions and stroke order. In contrast, handwriting with long, continuous strokes, as seen in Latin signatures (DeepSignDB), benefits more from frequency features. Frequency modeling extracts periodic patterns and harmonic components inherent in cursive writing styles, effectively capturing the overall shape and flow of pen movement. This adaptive gating demonstrates self-gated fusion's flexibility in tailoring feature emphasis to the unique characteristics of different handwriting styles, underscoring its interpretability and effectiveness.

\section{Visualization}
\label{appendix::visualization}
To more intuitively demonstrate the effectiveness of the temporal-frequency synergistic learning of SPECTRUM, we visualize the output feature sequence based on single-domain and multi-domain learning. Features are extracted from the same handwriting samples for comparison. We utilize the final output features of the Baseline model (as described in Sec.~\ref{sec::ablation}) for visualization in the temporal domain, while using the output features of the proposed SPECTRUM for visualization in the temporal-frequency domain. Visualizations are presented in Fig.~\ref{Fig::visualization}, which are performed on the Chinese signature data of MSDS-ChS and Token Digit String data of MSDS-TDS, respectively.

Comparing the left and right columns of each data type, the heatmaps on the right column showcase richer and denser regions with high response values, particularly evident in the Token Digit String data. This suggests that incorporating frequency features with temporal features strengthens the sensitivity of individual writing patterns, resulting in more informative handwriting representations and improved verification accuracy. In addition, as seen in the right-column heatmaps, the high-response regions are concentrated in areas such as stroke twirls, stroke hyphenations, and the start/end of strokes. These regions likely contain richer writing style characteristics, which are effectively captured by the frequency modeling approach. By highlighting these stylistically rich areas, our model demonstrates its ability to focus on crucial elements that distinguish individual writing patterns, further validating the effectiveness of our multi-domain approach.

\section{Limitation and Discussion}
\label{appendix::limitation}
Although SPECTRUM achieves optimal or SOTA-comparable performances on three datasets, the performance enhancement on Chinese/Latin signatures is less significant than on Token Digit String (TDS). This calls for further efforts to improve the generalizability of temporal-frequency learning on Chinese/Latin signatures. Additionally, our exploration of multi-domain learning has hitherto been confined to temporal and frequency domains. It is feasible to investigate other domains such as the spatial domain (rendering online data to offline images) and the video domain (capturing hand movements during writing), as well as the integration of more than two feature domains, to further enhance the robustness of online handwriting verification (OHV).

Furthermore, the successful integration of Chinese signature (ChS) and Token Digit String (TDS) indicates another simple yet effective avenue to improve OHV performance by combining multiple handwritten biometrics. This also offers potential benefits for real-world OHV applications, such as banking. Despite its straightforwardness, this approach remains unexplored, and available datasets are scarce. This underscores the need for further exploration in this area, such as using a broader range of handwritten biometrics beyond just ChS and TDS, collecting more comprehensive multi-biometric datasets, developing more effective techniques for biometric feature fusion, and integrating handwritten biometrics with other behavioral biometrics (\emph{e.g.}, face, fingerprint).

\end{document}